\documentclass{ecai} 

\usepackage{latexsym}
\usepackage{amssymb}
\usepackage{amsmath}
\usepackage{amsthm}
\usepackage{booktabs}
\usepackage{enumitem}
\usepackage{graphicx}
\usepackage{color}

\usepackage{algorithm}
\usepackage{wrapfig}
\usepackage[switch]{lineno}
\usepackage{xspace}
\usepackage{paralist}
\usepackage{todonotes}
\usepackage{algpseudocode}
\usepackage{comment}
\usepackage{subcaption}

\usepackage{multirow}
\usepackage{lscape}
\usepackage{longtable}

\usepackage{paralist}
\usepackage{afterpage} 

\newcommand{\triple}[3]{\ensuremath{\langle \texttt{#1}, \texttt{#2}, \texttt{#3} \rangle}}
\newcommand{\approach}{$\textsc{deCal}$} 
\newcommand{\keci}{\textsc{Keci}\xspace}

\newcommand{\BibTeX}{B\kern-.05em{\sc i\kern-.025em b}\kern-.08em\TeX}

\begin{document}


\begin{frontmatter}


\paperid{1574} 


\title{Embedding Knowledge Graphs in Degenerate Clifford Algebras}


\author{\fnms{Louis Mozart}~\snm{Kamdem Teyou}\orcid{0000-0001-7975-8794}}
\author{\fnms{Caglar}~\snm{Demir}\orcid{0000-0001-8970-3850}}
\author{\fnms{Axel-Cyrille}~\snm{Ngonga Ngomo}\orcid{0000-0001-7112-3516}} 

\address{Department of Computer Science, Paderborn University}



\begin{abstract}
Clifford algebras are a natural extension of division algebras, including real numbers, complex numbers, quaternions, and octonions. Previous research in knowledge graph embeddings has focused exclusively on Clifford algebras of a specific type, which do not include nilpotent base vectors—elements that square to zero. In this work, we introduce a novel approach by incorporating nilpotent base vectors with a nilpotency index of two, leading to a more general form of Clifford algebras named degenerate Clifford algebras. This generalization to degenerate Clifford algebras does allow for covering dual numbers and as such include translations and rotations models under the same generalization paradigm for the first time. We develop two models to determine the parameters that define the algebra: one using a greedy search and another predicting the parameters based on neural network embeddings of the input knowledge graph. Our evaluation on seven benchmark datasets demonstrates that this incorporation of nilpotent vectors enhances the quality of embeddings. Additionally, our method outperforms state-of-the-art approaches in terms of generalization, particularly regarding the mean reciprocal rank achieved on validation data. Finally, we show that even a simple greedy search can effectively discover optimal or near-optimal parameters for the algebra.
\end{abstract}

\end{frontmatter}


\section{Introduction}


Knowledge graphs (KGs) are used in an increasing number of applications and domains \cite{DBLP:journals/tkde/WangMWG17}. While several formalizations of KGs exist \cite{DBLP:journals/csur/HoganBCdMGKGNNN21}, we consider knowledge graphs $K \subseteq \mathcal{E}\times\mathcal{R}\times\mathcal{E} $, where $\mathcal{E}$ and $\mathcal{R}$ represent a set of entities and relations respectively. 
The elements of a knowledge graph are called \emph{assertions} (sometimes also facts), and are triples $\triple{x}{y}{z}$ where $\texttt{x}$ is called the head (also called subject), $\texttt{y}$ the relation (also called predicate) and $\texttt{z}$ the tail (also called object) of the assertion \cite{DBLP:journals/csur/HoganBCdMGKGNNN21}. 
For example, a KG may contain the triple \triple{Berlin}{capitalOf}{Germany}, which states that Berlin is the capital of Germany. This triple can be used to answer questions such as "What is the capital of Germany?" or "What is Berlin?" \cite{chen2020knowledge}. 
While knowledge graphs have existed since for decades \cite{DBLP:journals/csur/HoganBCdMGKGNNN21}, the term was popularized by Google in 2012 \cite{Singhal2012}, and its use has since surged \cite{chen2020knowledge,DBLP:journals/csur/HoganBCdMGKGNNN21}. 

 
KGs are often embedded into vector spaces to make them amenable to classical machine learning \cite{DBLP:journals/tkde/WangMWG17}. While initial approaches operated in $\mathbb{R}$ \cite{DBLP:journals/tnn/JiPCMY22},
it is evident from the existing literature that other division algebras facilitate the modelling of complex relations patterns, e.g., symmetry, and asymmetry. For example, the ability to model symmetric and asymmetric relations is conferred to ComplEx \cite{DBLP:conf/icml/TrouillonWRGB16} by its use of complex numbers. Recent embedding models have hence moved from real numbers to more complex number systems such as $\mathbb{C}$, $\mathbb{H}$, multi-vectors \cite{xu2020knowledge} and even Clifford Algebras $Cl_{p,q}(\mathbb{R})$ \cite{demir2023clifford}. 

In particular, 
embeddings in Clifford algebras $Cl_{p,q}(\mathbb{R})$ have recently been shown to achieve a significant improvement over the state of the art when used in combination with dimension scaling thanks to their ability to generalize over all normed division algebras  \cite{demir2023clifford} (see Table \ref{tab:clifford_subfields} for more details).  
The resulting approach, \keci, was shown to be a strict generalization of existing multiplicative embeddings approaches such as DistMult and ComplEx. However, none of the approaches based on dual numbers (e.g., \cite{DBLP:conf/aaai/CaoX0CH21}) could be generalized by this approach. This paper addresses this weakness by discarding the assumption of a non-degenerate quadratic form $Q$, which underpins \keci (see Section \ref{sec:preliminaries} for more details). In contrast, we assume that the quadratic form $Q$ that underpins our algebra can be degenerate, thus leading to $r$ base vectors being degenerate (e.g., nilpotent vectors $e_k$ with $e_k^2 = 0$). Our novel embedding approach, dubbed \approach\ (Embedding in \underline{de}generate \underline{C}lifford \underline{al}gebras), thus computes embeddings in $Cl_{p,q,r}(\mathbb{R})$. 


\section{Related Work}
Knowledge graph embeddings (KGE) models typically map an input KG $K$ into a low-dimensional and continuous vector space. According to the review in \cite{chen2020knowledge}, the plethora of KGE methods existing in the literature can be categorized into two main groups: translational models and multiplicative models. One of the first translational models is TransE \cite{bordes2013translating}: Given an assertion $\triple{x}{y}{z} \in K$, TransE's idea boils down to optimizing for $\mathbf{x}+\mathbf{y}\approx\mathbf{z}$ if \triple{x}{y}{z} holds, where $\mathbf{x},\mathbf{y},\mathbf{z}\in\mathbb{R}^d$ and represent the embeddings of $\texttt{x},\texttt{y}$ and $\texttt{z}$ respectively. The publication of TransE led to the development of several other similar models (e.g., TransH, TransR, TransD and RotatE) that address some of its main shortcomings, e.g., its poor modeling of reflexive, one-to-many, many-to-one and many-to-many relationships \cite{chen2020knowledge,ji2015knowledge,wang2014knowledge}. TransH \cite{wang2014knowledge} 
embeds knowledge graph by projecting entities and relations onto a hyperplane that is specific to each relation. Doing so allows TransH to effectively capture the mapping properties of relations, such as one-to-many and many-to-one, which TransE cannot handle \cite{chen2020knowledge,wang2014knowledge}. 
 TransR \cite{lin2015learning} introduces separate spaces for entities and relations, connected by a shared projection matrix. TransD \cite{ji2015knowledge} uses independent projection vectors for each entity and relation, which reduces the amount of computation compared to TransR. RotatE \cite{sun2019rotate} embeds the entities and relations into complex space and replaces the addition in TransE with the complex multiplication.

Multiplicative models like RESCAL, DistMult, ComplEx, QuatE, and OctE use bilinear transformations to score triples. RESCAL \cite{nickel2011three} employs a scoring function $\mathbf{x}^T\mathbf{A_y}\mathbf{z}$, where $\mathbf{x}$ and $\mathbf{z}$ are entity embeddings, and $\mathbf{A_y}$ is the relation matrix. DistMult \cite{yang2014embedding} simplifies this by using a diagonal relation matrix, $\mathbf{D_y} = \text{diag}(\mathbf{y})$, which is effective for symmetric relations but struggles with anti-symmetric ones. ComplEx \cite{trouillon2016complex} addresses this by embedding entities and relations in the complex space $\mathbb{C}$, using the real part for symmetry and the imaginary part for anti-symmetry. QuatE \cite{zhang2019quaternion} extends these capabilities to the quaternion space $\mathbb{H}$, allowing for the modeling of complex relationships such as inversion. OctE builds on QuatE by operating in the octonion space $\mathbb{O}$. However, both QuatE and OctE can face scaling challenges inherent to quaternion and octonion spaces. QMult and OMult \cite{demir2021convolutional} address this limitation through batch normalization, effectively mitigating the bottleneck.

In addition to translational and multiplicative models, the literature also features hyperbolic embedding methods such as RotH \cite{chami2020low} and MuRP \cite{balazevic2019multi}, which maps entities and relations from a knowledge graph onto a hyperbolic space, leveraging the properties of hyperbolic geometry to capture hierarchical structures in KGs. Dual quaternion methods like DualE \cite{cao2021dual} use dual quaternions \cite{jia2013dual} to embed entities and relations, offering a representation that combines the advantages of both hyperbolic and complex spaces. Rotational methods like RotE \cite{chami2020low} focus on learning rotation operations to capture relational patterns, while Euclidean methods such as MuRE \cite{balazevic2019multi}, which is a variant of MuRP, operate in Euclidean space, offering a simpler alternative for certain types of knowledge graphs.


\section{Motivation and Contribution}
The choice of the sub-algebra or space for embedding given any input knowledge graph plays a crucial role in computing an accurate representation of the input data as indicated in \cite{demir2023clifford}.
For instance, if a KG does not contain anti-symmetric relations, employing a complex-valued approach like ComplEx is likely to be less effective than using a simpler real-valued approach like DistMult. To address this, Demir et al. \cite{demir2023clifford} proposed incorporating the sub-algebra selection into the learning process by performing embeddings in Clifford Algebras $Cl_{p,q}(\mathbb{R})$ via the {\keci} model. Thanks to parameters $p$ and $q$, the {\keci} model is able to determine which space is appropriate to embed an input knowledge graph. As shown in Table \ref{tab:clifford_subfields}, {\keci} can decide or not if the embeddings will be carried out in $\mathbb{R}$, $\mathbb{C}$, $\mathbb{H}$, $\mathbb{O}$ and even beyond (for instance by setting $p=3$ and $q=4$).  

We build upon this idea via two main contributions:
\begin{compactitem}
\item We drop {\keci}'s assumption that the quadratic form underlying our Clifford algebra must not be degenerate and show how to embed even in degenerate Clifford algebras. Therewith, we can generalize over approaches based on dual numbers in addition to generalizing over \keci itself giving rise to more degree of freedom for our approach.

\item Moreover, we address the main weakness of dimension scaling: low dimension weights mean that particular dimensions barely contribute to the total score of \keci. Instead of learning $p$ and $q$ concurrently to the optimizing of the embeddings---hence effectively discarding dimensions without replacement---we present 2 approaches to predict $p$, $q$, and $r$ before we run \approach, and hence make full use of all dimensions available. 

\item Our implementation of \approach{} is publicly available to ensure that all results and experiments presented herein can be replicated. \footnote{https://github.com/dice-group/dice-embeddings}
\end{compactitem}
In the following sections, we present the technical details of {\approach} and provide evidence of its efficacy in link prediction of KG tasks.

\begin{table}[tb]
\centering
\caption{Relation between Clifford algebras and division algebras. N/A indicates no possible relationship between the spaces.}

\begin{tabular}{c c c c c}
\toprule
\textbf{Space} & $\subseteq$ & $Cl_{p,q}(\mathbb{R})$&  $\equiv$ & $Cl_{p,q,0}(\mathbb{R})$ \\ \midrule
$\mathbb{R}$ & &$Cl_{0,0}(\mathbb{R})$&  &$Cl_{0,0,0}(\mathbb{R})$\\ 
$\mathbb{C}$ & &$Cl_{0,1}(\mathbb{R})$ & &$Cl_{0,1,0}(\mathbb{R})$\\ 
$\mathbb{H}$ & &$Cl_{0,2}(\mathbb{R})$ & &$Cl_{0,2,0}(\mathbb{R})$\\ 
$\mathbb{O}$ & &$Cl_{1,3}(\mathbb{R})$ & &$Cl_{1,3,0}(\mathbb{R})$\\ 
$\mathbb{R}(\epsilon)$ & & N/A &  &$Cl_{0,0,1}(\mathbb{R})$\\ 
$\mathbb{R}^4(\epsilon)$ & & N/A &  &$Cl_{0,3,1}(\mathbb{R})$\\
\bottomrule
\end{tabular}

\label{tab:clifford_subfields}
\end{table}

\section{Clifford Algebras}
\label{sec:preliminaries}

\subsection{Definition}
A Clifford Algebra, denoted as $Cl(V,Q)$ \cite{clifford1882mathematical,clifford2010clifford}, is generated by a vector space $V$ and a quadratic form $Q$. When $V$ is a real vector space and $Q$ is a non-degenerate quadratic form, $Cl(V,Q)$ can be represented as $Cl_{p,q}(\mathbb{R})$ (where $p$ and $q$ are natural integers), signifying that $V$ possesses an orthogonal basis with $p+q$ vectors. Here, $p$ vectors $e_i$ satisfy $e_i^2 = 1$, and $q$ vectors $e_j$ satisfy $e_j^2 = -1$. The space $Cl_{p,q}(\mathbb{R})$ thereby extends traditional spaces such as the real and the complex space (see Table \ref{tab:clifford_subfields}).
In this paper, \emph{we refrain from assuming that the quadratic form $Q$ is non-degenerate}. Then, the orthogonal basis of our algebra consists of $p+q+r$ vectors, of which $p$ vectors (denoted $e_i$) are such that $e_i^2 = 1$, $q$ vectors (denoted $e_j$) are such that $e_j^2 = -1$, and $r$ vectors (denoted $e_k$) are such that $e_k^2 = 0$. Hence, the space $Cl(V,Q)$ is now denoted $Cl_{p,q,r}(\mathbb{R})$. Clearly, the algebra $Cl_{p,q,r}(\mathbb{R})$ is a super-algebra of  $Cl_{p,q}(\mathbb{R})$ for any $p, q, r \geq 0$.

\subsection{Norm  }

We consider $\mathbf{x}\in Cl_{p,q,r}(\mathbb{R}^d)$ that we represent using  $p+q+r$ orthogonal vectors as $\mathbf{x} = x_0 + \sum_{i=1}^{p}x_ie_i + \sum_{j=p+1}^{p+q}x_je_j + \sum_{k=p+q+1}^{p+q+r}x_ke_k$, where $x_{(\cdot)}\in\mathbb{R}^{\lfloor \frac{d}{1+p+q+r}\rfloor}$.  
The norm of ${\mathbf{x}}$, denoted $\|\mathbf{x}\|$, is defined using the quadratic form $Q$ that govern $Cl_{p,q,r}$. Since $Q$ is degenerate, the norm only operates on non-degenerate vectors \cite{keller2012structure} , i.e. 
\begin{equation}
    \|\mathbf{x}\|^2 = {Q(\mathbf{x})} =x_0^2 + \sum_{i=1}^{p}x_i^2e_i^2- \sum_{j=p+1}^{p+q}x_j^2e_j^2  = \sum_{i=0}^{p+q}x_i^2.
\end{equation}

\subsection{Inner Product and Clifford Product}
If we take also  $\mathbf{y}\in Cl_{p,q,r}(\mathbb{R}^d)$ such that \begin{equation}
    \mathbf{y} = y_0 + \sum_{i=1}^{p}y_ie_i + \sum_{j=p+1}^{p+q}y_je_j + \sum_{k=p+q+1}^{p+q+r}y_ke_k,
\end{equation}
the inner product $\mathbf{x}\cdot\mathbf{y}$ between ${\mathbf{x}}$  and ${\mathbf{y}}$ is given by
\begin{align*}
      x_0\cdot y_0 + \sum_{i=1}^{p}x_i\cdot y_i +\sum_{j=p+1}^{p+q}x_j\cdot y_j +\sum_{k=p+q+1}^{p+q+r}x_k\cdot y_k.
\end{align*} 
The Clifford product $ {\mathbf{x}}\circ{\mathbf{y}}$ between vectors ${\mathbf{x}}$ and ${\mathbf{y}}$ involves element-wise multiplication of their components across different basis elements (see supplementary material \cite{kamdem_teyou_2024_13341499} for mathematical details).

\section{Approach} 
\label{Approach}

To ensure comparability with standard approaches, we address the embedding problem in a $d$-dimensional vector space. Consequently, DistMult performs embeddings into $\mathbb{R}^d$, ComplEx into $\mathbb{C}^{d/2}$, QMult (which has demonstrated slightly better results than QuatE) and OMult (which has shown slightly better results than OctE) \cite{demir2021convolutional} into $\mathbb{H}^{d/4}$ and $\mathbb{O}^{d/8}$ respectively ). Additionally, {\keci} embeds into $Cl_{p,q}(\mathbb{R}^{m'})$, where $m' = \lfloor\frac{d}{1+p+q}\rfloor$, and {\approach} into $Cl_{p,q,r}(\mathbb{R}^m)$, with $m = \lfloor\frac{d}{1+p+q +r}\rfloor$ (see Table \ref{tab:comp_time}).

\subsection{Embedding in Degenerate Clifford Algebras}
Considering a triple $\triple{x}{y}{z} \in K$, we represent the embeddings  $\mathbf{x}$ and $\mathbf{y}$ of $\texttt{x}$ and $\texttt{y}$  in the space $Cl_{p,q,r}(\mathbb{R}^m)$ as:
\begin{align}
    &\mathbf{x} = x_0 + \sum_{i=1}^{p}x_ie_i + \sum_{j=p+1}^{p+q}x_je_j +\sum_{j=p+q+1}^{p+q+r}x_ke_k\\
    &\mathbf{y} = y_0 + \sum_{i=1}^{p}y_ie_i + \sum_{j=p+1}^{p+q}y_je_j +\sum_{j=p+q+1}^{p+q+r}y_ke_k,
\end{align}
with $x_{(\cdot)}, y_{(\cdot)} \in \mathbb{R}^m $
and 
\begin{align}
\label{Eq:cond}
    \begin{cases}
    e_i^2 = 1,\ \forall i \in \{1,\cdots,p\}\\
    e_j^2 = -1,\ \forall j \in \{p+1,\cdots,p+q\}\\
    e_k^2 = 0,   \ \forall k \in \{p+q+1,\cdots,p+q+r\}\\
    e_{\ell}e_n = - e_ne_{\ell}, \ \forall n\neq \ell.
    \end{cases}
\end{align}
Using Equations \ref{Eq:cond} and adopting the notations from \cite{demir2023clifford} the Clifford multiplication between the head and relation representation is:
\begin{align*}
\mathbf{x}\circ \mathbf{y} = \ &\sigma_{0}+\sigma_{p}+\sigma_{q}+\sigma_{r}+\sigma_{p,p}+\sigma_{q,q}+\sigma_{r,r}+\sigma_{p,q}+\\
&\sigma_{p,r}+\sigma_{q,r}.
\end{align*}
The terms $\sigma_0$, $\sigma_p$, $\sigma_q$, $\sigma_{p,p}$, $\sigma_{q,q}$, and $\sigma_{p,q}$ are as in \cite{demir2023clifford}, with new terms derived in supplementary material \cite{kamdem_teyou_2024_13341499} and defined as follows:
\begin{align}
  & \sigma_r = \sum_{k=p+q+1}^{p+q+r}\left(x_0y_k+x_ky_0\right)e_k\\
    &\sigma_{r,r} = \sum_{k=p+q+1}^{p+q+r-1}\sum_{k'=k+1}^{p}(x_ky_{k'}-x_{k'}y_k)e_ke_{k'}\\
    &\sigma_{p,r} = \sum_{i=1}^{p}\sum_{k=p+q+1}^{p+q+r}(x_iy_k - x_ky_i)e_ie_k\\
   & \sigma_{q,r} = \sum_{j=p+1}^{p+q}\sum_{j=p+q+1}^{p+q+r}(x_jy_k - x_ky_j)e_je_k.
\end{align}
\subsection{Scoring Function Derivation}
The scoring function of  {\approach} consists of taking the Clifford multiplication between the embeddings of the head and the relation, followed by a scalar product with the tail embeddings $\mathbf{z}$ i.e. \begin{align}
 \text{\approach}\big(\triple{x}{y}{z}\big)= (\mathbf{x}\circ\mathbf{y})\cdot\mathbf{z}.
\end{align}
Since the Clifford multiplication $\mathbf{x}\circ\mathbf{y}$ generate multi-vectors coordinates, we represent $\mathbf{z} $ in order to perform the above scalar product as \begin{align} 
     \mathbf{z}=\ &z_0 + \sum_{i=1}^{p}z_ie_i + \sum_{j=p+1}^{p+q}z_je_j +\sum_{j=p+q+1}^{p+q+r}z_ke_k + \Vec{Cte}
\end{align}

where $z_{(\cdot)}\in\mathbb{R}^m$ and  $\Vec{Cte}$ represent a unitary multi-vector \cite{kamdem_teyou_2024_13341499}.
That is, in the tail representation, all coefficients of the multi-vectors involved in the scalar product are set to one. Hence, the scoring function of {\approach} can be deduced from the score of {\keci} as,
\begin{align*}
    \text{ \approach}\big(\triple{x}{y}{z}\big) = \ & \text{\keci}\big(\triple{x}{y}{z}\big)+\sigma_{r,r}^*+\sigma_{p,r}^*+\sigma_{q,r}^*\\
&+\sum_{k=p+q+1}^{p+q+r}\left(x_0y_kz_k+x_ky_0z_k\right),
\end{align*}
where
$\sigma_{r,r}^*,\sigma_{p,r}^*$ and $\sigma_{q,r}^*$ represents the sum of the coordinates of the multi-vectors $\sigma_{r,r},\sigma_{p,r}$ and $\sigma_{q,r}$ respectively.


\section{Embedding Space Search}
\label{sec:spacesearch}
Finding adequate values for $p$, $q$, and $r$ is bound to be of central importance when embedding using \approach. 
An exhaustive search over these parameters results in $\frac{1}{6}(d+1)(d+2)(d+3) \in O(d^3)$ possible combinations,  
 where $d$ represents the embedding dimension. For instance, with $d=16$, the model would need to be run 969 times, proving an inefficient and time-consuming approach. 
 To address this challenge, we developed four strategies to navigate the parameter space defined by $p$, $q$, and $r$. While two of these serve as baseline approaches, the other two represent our key contributions, focusing on finding configurations that yield the highest mean reciprocal rank on the validation data by learning from the training set and applying this combination on the test set.

\paragraph{Local Exhaustive Search (LES)}

The local exhaustive search involves systematically exploring all potential parameter (without constraint) combinations for {\approach} in a subspace of the parameter space $\{0, 1,\cdots,d\}^3$ with $0 \leq p+q+r \leq d$. 

\paragraph{Global Search with Divisibility Criterion (GSDC)}
This approach subsamples the space covered by the local exhaustive search by only visiting spaces where $(1+p+q+r)$ divides $d$. The motivation behind this approach is to only consider configuration which fully exploits the total number of dimensions available to \approach. For $d=16$, the approach visits 186 Clifford algebras.


\paragraph{Greedy Search (GS)}
Our first approach to optimize the hyperparameters $p$, $q$ and $r$ of {\approach} consists of the greedy search algorithm described in Algorithm \ref{AlgoGBS}. The Algorithm starts with the initial configuration $(p,q,r)=(1,1,1)$ and iteratively generates new configurations (unseen configurations) in a local neighbourhood by adding 1, -1 or 0 to $p$, $q$, and $r$. Then, we evaluate the mean reciprocal rank (short: MRR) for each unseen configuration and append these scores to a queue which is then sorted in descending order based on the MRR. The next configuration to evaluate is then selected from the highest-scoring configurations. This process is repeated until convergence or the maximum number of iterations is reached. The algorithm terminates if the best configuration remains unchanged, indicating a local maximum.

\begin{algorithm}[tb]
  \caption{GreedySearch}  
 \begin{algorithmic}[1]
    \Function{Optimal\_params}{$\text{max\_iterations}$}
      \State $p_0, q_0, r_0 \gets 1, 1, 1$ \Comment{Initialization}
      \State $\text{seen\_conf} \gets []$
      \State $\text{prior\_query} \gets []$
      \For{$i \in [0, \text{max\_iterations})$}
        \State $\text{to\_score} \gets \textsc{GenerateConf}(\text{seen\_conf}, p, q, r)$ 
        \State $\text{prior\_query} \gets \textsc{Score}(\text{to\_score}, \text{prior\_query})$ 
        \State $\text{prior\_queue} \gets \text{sort}(\text{prior\_query})$ 
        \State $p, q, r, \text{max\_MRR} \gets \text{prior\_queue}[0]$
        \If{$(p,q,r)=(p_0,q_0,r_0)$} 
          \State \textbf{break} \Comment{Local maximum found}
        \Else
          \State $p_0, q_0, r_0 \gets p, q, r$ \Comment{Update parameters}
          \State $\text{seen\_conf} = \text{seen\_conf}\cup\text{to\_score}$
        \EndIf
      \EndFor
      \State \Return $(p, q, r, \text{max\_MRR})$
    \EndFunction
       \Function{GenerateConf}{$\text{queue}, p, q, r$}
      \State $\ell \gets []$
      \For{$p_i,q_i,r_i \in [-1, 0, 1]$}
            \If{$(p+p_i, q+q_i, r+r_i) \notin \text{queue}$}
              \State $\ell = \ell \cup  \{(p+p_i, q+q_i, r+r_i)\} $
            \EndIf
      \EndFor
      \State \Return $\ell$
    \EndFunction
    
    \Function{Score}{$\text{queue, prior}$}
      \For{$(p,q,r) \in \text{queue}$}
        \If{$(p \geq 0) \land (q \geq 0) \land (r \geq 0)$}
          \State $\text{prior} = \text{prior} \cup \{(p, q, r, \text{MRR}(p,q,r))\}$
        \EndIf
      \EndFor
      \State \Return $\text{prior\_queue}$
    \EndFunction



    \Function{MRR}{$p,q,r$}
      \State \Return {\approach}.MRR(p,q,r) \Comment{MRR Evaluation}
    \EndFunction
  \end{algorithmic}
\label{AlgoGBS}
\end{algorithm}

\begin{figure}[tb]
    \centering
    \subcaptionbox{KINSHIP}{\includegraphics[width=0.4\linewidth]{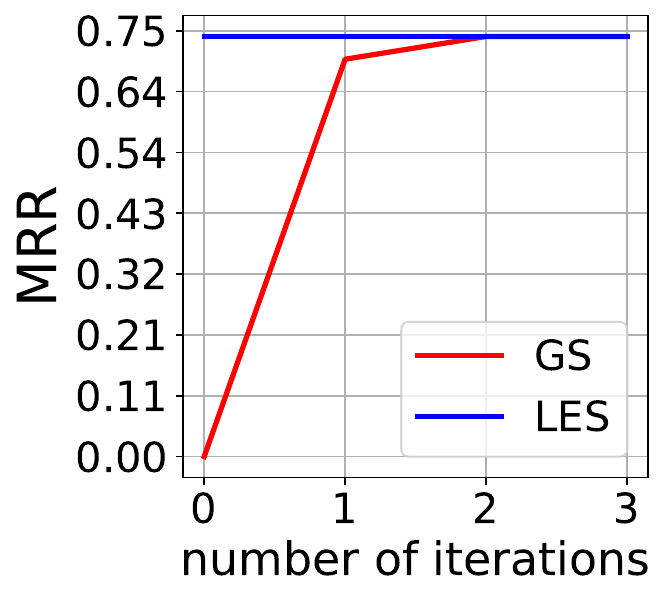}}
    \subcaptionbox{UMLS}{\includegraphics[width=0.4\linewidth]{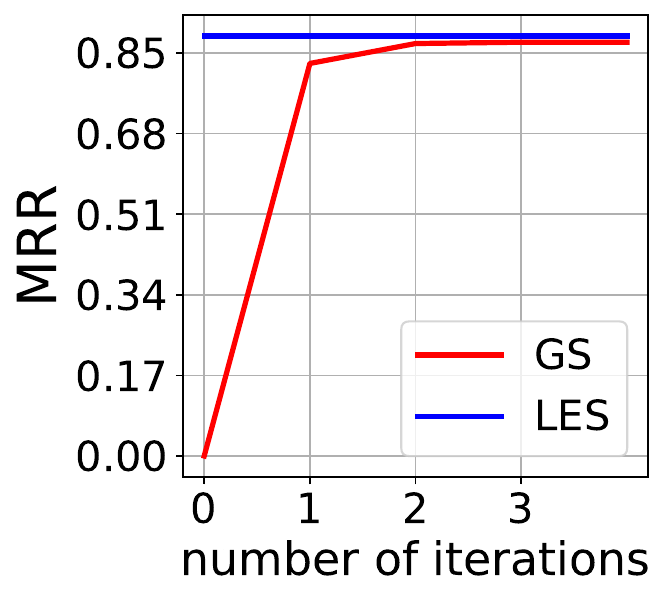}}
    \vspace{.5cm}
    
    \subcaptionbox{NELL-995-h100}{\includegraphics[width=0.4\linewidth]{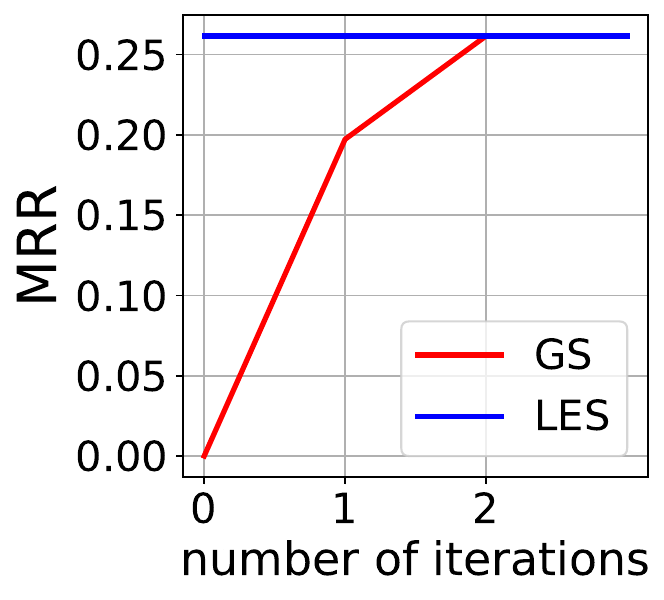}}
    \subcaptionbox{NELL-995-h75}{\includegraphics[width=0.4\linewidth]{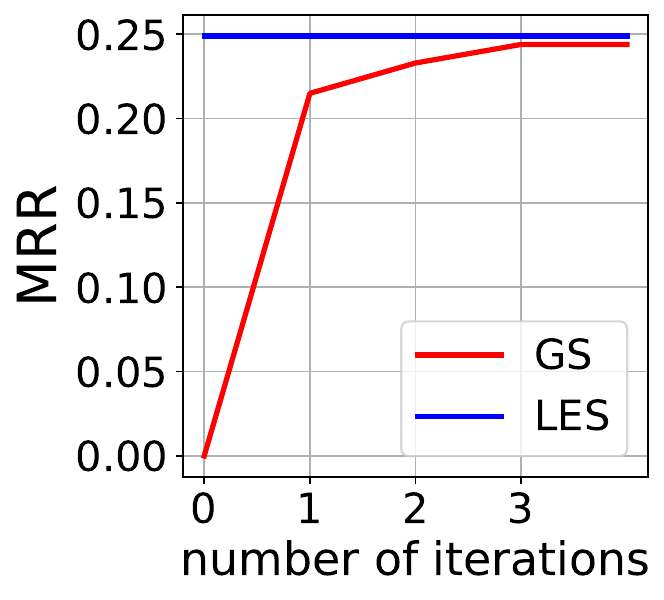}}
    \vspace{.5cm}
    
    \subcaptionbox{NELL-995-h50}{\includegraphics[width=0.4\linewidth]{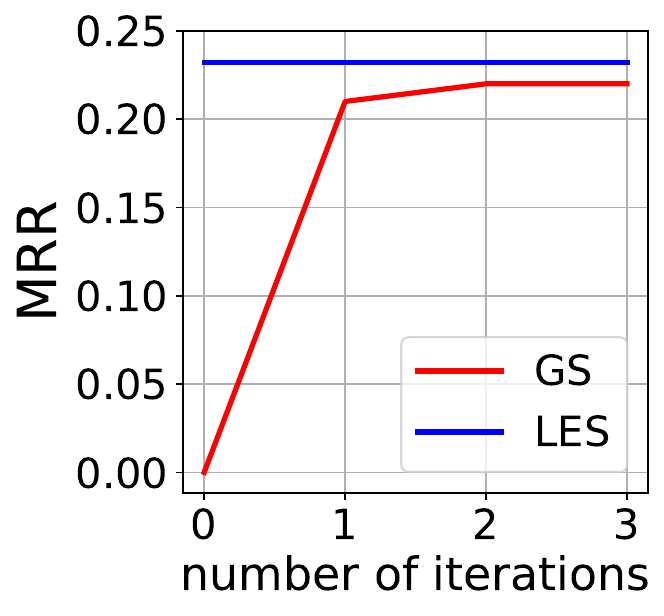}}
    \subcaptionbox{FB15k-237}{\includegraphics[width=0.4\linewidth]{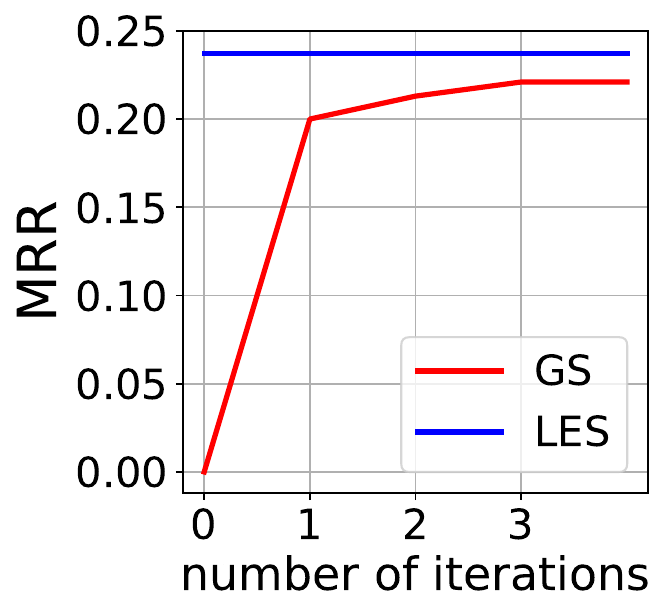}}
    \vspace{.5cm}
    
    \subcaptionbox{WN18-RR}{\includegraphics[width=0.4\linewidth]{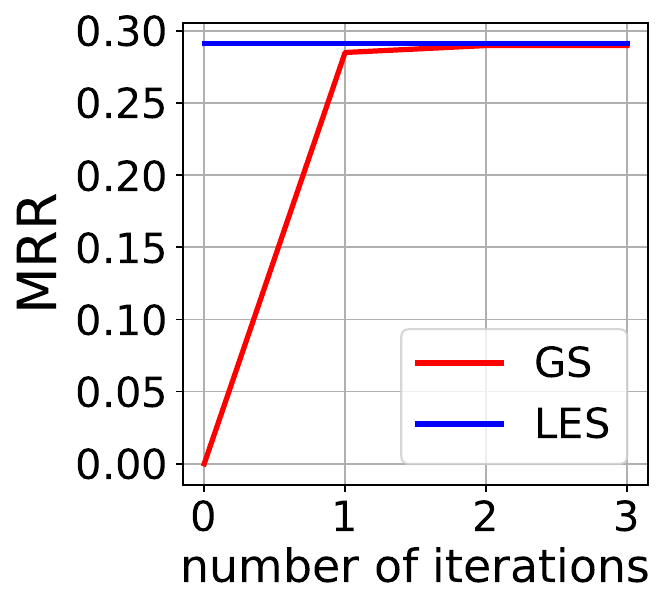}}
    \vspace{.5cm}
    \caption{GS convergence speed to LES across benchmark datasets.}
    \vspace{.5cm}
    \label{fig:subfigures}
\end{figure}

A major limitation of this approach is that its can only detect local maxima close to its starting point. While it performs well in practice (see Section \ref{sec:experiments}), we devise another approach able to optimize $p$, $q$, and $r$ using global information. 

\paragraph{Vector Space Prediction (VSP)}
Given an input knowledge graph $K$, this approach aims to predict the optimal values of $p$, $q$, and $r$ for $K$ w.r.t. the MRR. To achieve this goal, our approach  begins by computing an embedding of $K$ in $Cl_{1, 1, 1}$. The embedding of each triple $\triple{x}{y}{z} \in K$ is the concatenation of the embedding of $\texttt{x}$, $\texttt{y}$ and $\texttt{z}$. The embeddings of all triples in $K$ are finally used as input for a pre-trained recurrent neural network, which predicts the values of $p$, $q$ and $r$ for $K$. 





\section{Results and Discussion}
\label{sec:experiments}
\subsection{Datasets and Experimental Setup}
We evaluate our model on seven benchmark datasets, comprising five large datasets (WN18-RR, FB15k-237, NELL-995-h100, NELL-995-h50, and NELL-995-h75) and two smaller datasets (UMLS, and KINSHIP). For detailed statistics on these datasets, please refer to Table \ref{data_set_stats}.


\begin{table}[htb]
    \centering
    \caption{Overview of benchmark datasets.}
    \small
     \resizebox{\linewidth}{!}{
    \begin{tabular}{cccccc}
    \toprule
    \textbf{Dataset} & \multicolumn{1}{c}{$|\mathcal{E}|$}&  $|\mathcal{R}|$ & $|\mathcal{G}^{\text{Train}}|$ & $|\mathcal{G}^{\text{Validation}}|$ &  $|\mathcal{G}^{\text{Test}}|$\\
    \hline
    \midrule
    WN18-RR        &40,943      &22  &86,835       &3,034&3,134\\
    FB15k-237           & 14,541     &237 & 272,115       & 17,535&20,466\\ 
    NELL-995-h50        & 34,667      &86  & 72,767       & 5,440& 5,393\\
    NELL-995-h75           & 28,085    & 114  &59,135     &4,441&4,389\\ 
    NELL-995-h100       &22,411      & 86  & 50,314      &3,763&3,746\\
    \midrule
    \midrule
    UMLS           &135      &46  &5,216       &652&661\\ 
    KINSHIP        &104      &25  &8,544       &1,068&1,074\\
    \bottomrule
    \end{tabular}}
     
     \label{data_set_stats}
\end{table}

We conducted two series of experiments. First, we wanted to quantify how the different approaches for determining $p$, $q$ and $r$ performed. To this end, we evaluated the combination of \approach with the approaches presented in Section \ref{sec:spacesearch}. 
For the vector space prediction (VSP), we used a leave-one-out training setting, where we used 1000 subgraphs containing 5000 triples from 6 benchmark datasets for training and the remaining dataset for testing. We trained three distinct models: LSTM (Long Short-Term Memory), GRU (Gated Recurrent Unit), and a concatenation of both \cite{fu2016using}.  

\begin{table}[tb]
\centering
\caption{Embedding spaces and model time complexity comparison. GSDC serves as the reference method, with relative times indicating the efficiency of each model compared to GSDC. Note that the recorded time of VSP reflects only its prediction phase. }
\label{tab:comp_time}
\begin{tabular}{lcc}
\toprule
\textbf{Models}   & \textbf{Embedding Space}               & \textbf{Relative Time}\footnote{(\% of Best Method)} \\ \midrule
GSDC     & \multirow{4}{*}{$Cl_{p,q,r}(\mathbb{R}^m)$,\ $m=\lfloor\frac{d}{1+p+q+r}\rfloor$} & 100 \%                            \\
LES      &                     & 67.6\%                              \\
GS       &                     &  24.3\%                              \\
VSP      &                     &   5.2\%                                \\ 

\midrule
Keci     & $Cl_{p,q}(\mathbb{R}^m)$, \ $m=\lfloor\frac{d}{1+p+q}\rfloor$               &     56.7\%                              \\ 
\midrule

DistMult & \multirow{4}{*}{$\mathbb{R}^d$}  &    10.1\%                              \\
RotE     &                     &            12.5\%                       \\
MuRE     &                     &          20.2\%                         \\
MuRP     &                     &           22.4\%                        \\ \midrule
CompEx   & $\mathbb{C}^{d/2}$                   &    10.9\%                               \\ \midrule
QMult    & $\mathbb{H}^{d/4}$                   &     11.4\%                              \\ \midrule
DualE    & $\mathbb{R}^d(\epsilon)$                    &    12.5\%                               \\ \midrule
OMult    & $\mathbb{O}^{d/8}$                   &              11.6\%                     \\ \bottomrule
\end{tabular}
\end{table}

In our second series of experiments, we conducted a comparison between \approach{} and state-of-the-art algorithms. we reported the performance of each algorithm on the test data, and for results on train and validation data, refer to the supplementary material \cite{kamdem_teyou_2024_13341499}.
We used the same set of parameters for all approaches: $d =16$, number of epochs = 250, batch size = 1024, learning rate = 0.1, Adam optimizer and the KvsAll training technique.
Like previous works \cite{DBLP:journals/tkde/WangMWG17,demir2023clifford}, we used the Mean Reciprocal Rank (MRR) and hits at 1, 3, and 10 as performance measures. All experiments presented in this paper were conducted on a virtual machine equipped with two NVIDIA H100-80C GPUs, each with 80 GB of memory.

\subsection{Comparison of Different Variants}

We evaluated the performance of {\approach} in combination with different strategies for discovering $p$, $q$, and $r$. 
An overview of these results can be found in the four bottom lines of  Tables~\ref{tab:link_pred2} and ~\ref{tab:link_pred3}. The found Clifford space as well as the computational time of the different search algorithms of \approach{} and \keci{} on all data sets can be found in Table 1 of the supplementary material \cite{kamdem_teyou_2024_13341499}.
The combinations of {\approach} with LES and GSDC can be regarded as upper bounds of the performance of our algorithm. 
Our approach outperformed all other variants (as well as the state of the art) when combined with GSDC except on UMLS, where {\approach} + LES performed best.   
We conclude that GSDC is the approach of choice when aiming to determine useful values for $p$, $q$ and $r$.  
However, the number of combinations of these parameters that need to be checked can be prohibitively large, especially if $d$ is a highly composite number. 

In our experiments, using GS was less time-consuming than GSDC (see Table \ref{tab:comp_time}) but led to slightly worse results w.r.t. the MRR on the benchmark datasets. 
For example, GS outperformed GSDC on the UMLS dataset, achieved the same performance as GSDC on KINSHIP, and reached over 91.6\% of GSDC's performance on average.
This suggests that GS can indeed be used for configuring {\approach} if the number of combinations of $p$, $q$, $r$ to explore is to be kept low.
Note that GS only needed at most three iterations in our experiments to find a local maximum (see Figure \ref{fig:subfigures}).
This strength of GS is also its main weakness as the convergence of this method depends significantly on the initial starting point. The MRR function as shown in Figure 1 in the supplementary material \cite{kamdem_teyou_2024_13341499}, shows multiple local maxima, making the search for the global maximum challenging. 
Still, our results indicate that local maxima generally suffice to achieve a state-of-the-art performance. 
 
\begin{figure}[htb]
    \centering
    \includegraphics[width=.8\linewidth]{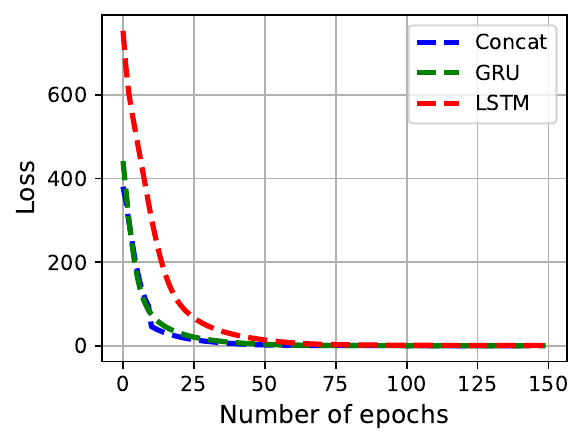}
    \caption{Loss function curves.}
    \vspace{.5cm}
    \label{figloss}
\end{figure}
Figure \ref{figloss} illustrates the loss evolution during the training phase for each model on all datasets with FB15k-237 left out.  On the test data, the LSTM, GRU, and concatenated models achieved a prediction accuracy of 44.5\%, 42.0\%, and 35.5\%, respectively, where a prediction was considered correct if it returned the $(p, q, r)$ combination suggested by LES. Employing an ensemble approach with weights of 0.75, 0.2, and 0.05 for the LSTM, GRU, and concatenated models, our predictor achieved a superior performance of 45.0\%.  
The combination of the concatenated model of VSP and {\approach} turned out to achieve approximately 84.5\% of the MRR of {\approach} + GDSC on average. The results are not surprising given that the approach was trained on a small number of samples. We hypothesize that the performance of the approach can be much improved with the availability of more diverse training data. Still, the leave-one-out strategy we employed for learning suggests that a universal predictor can indeed be trained to predict suitable values of the Clifford algebra parameters. This will be the subject of future works.

\subsection{Comparison with other Approaches}

\begin{table}[tb]
    \centering
    \caption{Link prediction results on WN18-RR at the test time. Bold and underlined results indicate the best and second-best results respectively.}
    \small
    \begin{tabular}{l cccc}
    \toprule
    \textbf{Models} & \multicolumn{4}{c}{\textbf{WN18-RR}} \\
    \cmidrule(l){2-5}
    & MRR & H@1 & H@3 & H@10 \\
    \midrule
    DistMult 
             & 0.231 & 0.191& 0.249 & 0.305 \\
    \midrule
    ComplEx 
             & 0.274 & 0.233 & 0.295 & 0.349 \\
    \midrule
    QMult    
             & 0.242 & 0.199 & 0.262 & 0.318 \\
    \midrule
    OMult    
             & 0.121 & 0.080 & 0.135 & 0.203 \\

     \midrule

    MuRE& 0.259 & 0.202 & 0.280 & 0.366 \\

    \midrule

    MuRP& 0.212 & 0.160 & 0.234 & 0.309 \\
    
    \midrule
    
    RotH& 0.230 & 0.177 & 0.247 & 0.332 \\
    
    \midrule
    
    RotE& 0.258 & 0.203 & 0.280 & 0.357 \\
    
    \midrule

    {DualE}    
             & 0.157 & 0.118 & 0.174 & 0.228 \\
    \midrule
   {\keci}    
             & 0.285 & 0.245 & 0.302 & 0.357 \\

    \midrule
   
    \midrule
    
 {\approach} + LES
                 & 0.291 & 0.249 & 0.313 & 0.364\\
    {\approach} + GSDC  
             & \textbf{0.296} & \underline{0.250} & \textbf{0.323} & \textbf{0.381} \\
  {\approach} + GS  
                   & \underline{0.295} & \textbf{0.252} &\underline{ 0.317} & \underline{0.372}\\
  {\approach} + VSP   
                 & 0.285 & 0.245 & 0.302 & 0.357\\
    \bottomrule
    \end{tabular}
    
    \label{tab:link_pred3}

\end{table}

\begin{table*}[htb]
    \centering
    \caption{Link prediction results on  UMLS, KINSHIP, NELL-995-h100, NELL-995-h75, NELL-995-h50 and FB15k-237 on the test data. Bold and underlined results indicate the best and second-best results respectively.}
    \small
\begin{tabular}{@{}l cccc ccccc ccccc cccc c@{}}
    \toprule
     \textbf{Models} &\multicolumn{4}{c}{\textbf{UMLS}} && \multicolumn{4}{c}{\textbf{KINSHIP}}&& \multicolumn{4}{c}{\textbf{NELL-995-h100}}\\
     \cmidrule(l){2-5} \cmidrule(l){7-10} \cmidrule(l){12-15} 
        &MRR  &H@1   &H@3   &H@10 && MRR & H@1  & H@3  & H@10  && MRR & H@1  & H@3  & H@10\\
    \midrule
DistMult 

& 0.756 & 0.647 & 0.834 & 0.952 && 0.514 & 0.346 & 0.606 & 0.878 && 0.185 & 0.120 & 0.205 & 0.320\\
\midrule
ComplEx 
&  0.835 & 0.728 & 0.939 & 0.976 &&\underline{0.725} &\underline{0.594 }&\underline{0.821} &  {0.963}&& 0.168 & 0.109 & 0.187 & 0.285\\

\midrule
QMult  
&0.859 & 0.773 & 0.936 & 0.980 &&0.624 &0.481 &0.715 & 0.911&& 0.125 & 0.079 & 0.135 & 0.214\\

\midrule

OMult&0.827 & 0.756 & 0.874 & 0.954 &&0.488 &0.365 &0.536& 0.747&& 0.137 & 0.088 &  0.151 & 0.237\\

\midrule

MuRE& 0.876 & 0.782 & \textbf{0.964} & 0.990 && 0.664 & 0.526 & 0.754 & 0.933 && 0.236 & 0.167 &  0.261 & 0.374\\

\midrule

MuRP& 0.874 & \textbf{0.824} & 0.908 & 0.970 && 0.676 & 0.532 & 0.779 & 0.946 && 0.237 & 0.161 &  0.261 & 0.395\\

\midrule

RotH& 0.802 & 0.689 & 0.899 & 0.981 && 0.592 & 0.427 & 0.689 & 0.932 && 0.192 & 0.132 &  0.214 & 0.310\\

\midrule

RotE& 0.866 & 0.766 & \underline{0.962} &\textbf{ 0.994} && 0.709 & 0.569 & 0.814 & \textbf{0.967} && 0.225 & 0.157 &  0.250 & 0.361\\

\midrule

{DualE} 
&{0.866} &{ 0.774} & {0.955} & {0.985} &&{0.591} &{0.443} &{ 0.675}& {0.905}&& {0.173} & {0.113} &{ 0.188} & {0.295}\\

\midrule
{\keci } 
&{0.875} & 0.798 & {0.944} & {0.983} && \textbf{0.743}&\textbf{0.621} &  \textbf{0.830} &   \underline{0.964}&& {0.252} & {0.174} & \underline{0.288} & \underline{0.408}\\

\midrule

{\approach} + LES 
&\textbf{ 0.883} & \underline{0.799} & \underline{0.962 }& \underline{0.991 }&&  \textbf{0.743}&\textbf{0.621} &  \textbf{0.830} &   \underline{0.964}&&  \underline{0.256} & \underline{0.184} & {0.280}& {0.399}\\ 
{\approach} + GSDC
& {0.878} & \underline{0.799} &{0.947}& {0.989} &  &  \textbf{0.743}&\textbf{0.621} &  \textbf{0.830} &     \underline{0.964}& & \textbf{0.270} &\textbf{ 0.196} & \textbf{0.299} &\textbf{0.417} \\ 
{\approach} + GS
               & \underline{0.879} &{0.797} & {0.951 }& {0.987}&&  \textbf{0.743}&\textbf{0.621} &  \textbf{0.830} &     \underline{0.964}&& \underline{0.256} & \underline{0.184} & {0.280}& {0.399}\\
{\approach} + VSP  
                &{0.843} & {0.732} & {0.944} &{ 0.986}&&  {0.674}&{0.540} & {0.766} &   {0.942}&& 0.235 & 0.167 & 0.259 & 0.374\\ 

    \toprule
    \toprule


  \textbf{Models}    &\multicolumn{4}{c}{\textbf{NELL-995-h75}} && \multicolumn{4}{c}{\textbf{NELL-995-h50}}&& \multicolumn{4}{c}{\textbf{FB15k-237}}\\
     \cmidrule(l){2-5} \cmidrule(l){7-10} \cmidrule(l){12-15} 
        &MRR  &H@1   &H@3   &H@10                                       && MRR &H@1   &H@3   &H@10 && MRR &H@1   &H@3   &H@10\\
    \midrule
DistMult
& 0.163 & 0.104 & 0.179 &  0.282 && 0.151 &0.097 &0.168 & 0.256&&0.147 & 0.093 & 0.158 &  0.260\\
\midrule
ComplEx
&  0.139 &  0.085 & 0.156 &  0.245 && {0.163} & {0.107} &{0.183} &{ 0.275}&&  0.137 &  0.087 &  0.147 & 0.239\\
\midrule
QMult  
&0.158 & 0.106 & 0.171 & 0.261 &&  0.101 &0.0596 & 0.114 &  0.185&&  0.122 &  0.077 & 0.128 & 0.208\\
\midrule

OMult 
& 0.129 & 0.082 & 0.142 &  0.223 &&0.119 &0.076 &0.131 & 0.210&& 0.086 & 0.059 & 0.089 &  0.138\\
\midrule

MuRE& 0.229 & 0.163 & 0.256 & 0.357 && 0.215 & 0.151 & 0.237 & 0.345 && 0.209 & 0.147 &  0.224 & 0.333\\

\midrule

MuRP& 0.230 & 0.161 & 0.255 & 0.370 && 0.217 & 0.151 & 0.239 & 0.348 && 0.204 & 0.141 &  0.220 & 0.329\\

\midrule

RotH& 0.163 & 0.104 & 0.184 & 0.279 && 0.154 & 0.098 & 0.169 & 0.270 && 0.144 & 0.097 &  0.151 & 0.233\\

\midrule

RotE& 0.208 & 0.144 & 0.230 & 0.335 && 0.218 & 0.152 & 0.243 & 0.349 && 0.197 & 0.133&  0.211 & 0.324\\

\midrule
{DualE} 
&{0.178} &{0.120} & {0.196} & {0.291} &&{0.176} &{0.116} &{0.195}& {0.298}&& {0.185} & {0.125} &{ 0.199} & {0.299}\\

\midrule

{\keci} 
&{0.237} &{ 0.172} & {0.260} & {0.365} &&\textbf{0.250} &\textbf{0.177} &\textbf{ 0.281}& \textbf{0.392}&& \underline{0.235} & \underline{0.164} &{ 0.255} & \underline{0.376}\\

\midrule

{\approach} + LES 
                &\underline{0.245}&\underline{0.177}&\underline{0.274}&{0.374}&&\underline{0.232}&\underline{0.166}&\underline{0.253}&\underline{0.368}&&\underline{0.235}&\underline{0.164}&\underline{0.258}&0.375\\
{\approach} + GSDC
& \textbf{ 0.251} & \textbf{0.178 }& \textbf{ 0.281} &  \textbf{0.396}& &\textbf{0.250} &\textbf{0.177} & \textbf{0.281} &\textbf{0.392} && \textbf{0.241} & \textbf{0.171} & \textbf{0.263} &\textbf{0.380}\\
{\approach} + GS &0.230&0.170&0.268&\underline{0.376}&&0.163&0.106&0.183&0.273&&0.217&0.150&0.236&0.349\\
{\approach} + VSP &{0.202}&{0.136}&{0.225}&{0.335}&&0.203&0.138&0.229&0.333&&0.144&0.091&0.156&0.251\\

    \bottomrule
    \end{tabular}
    \label{tab:link_pred2} 
\end{table*} 

Tables~\ref{tab:link_pred2} and ~\ref{tab:link_pred3} present link prediction results on the datasets presented in Table \ref{data_set_stats}.  
In the following, we mainly focus on the performance of \approach{} + GSDC and \approach{} + GS when comparing our approach with the state of the art. 

The prediction results on the WN18RR datasets are shown in Table~\ref{tab:link_pred3}. The findings indicate that all variations of our approach outperformed the state-of-the-art model in all metrics, except for \approach{} + VSP, which predicted the same space for embeddings as \keci{}, resulting in similar performance.

The prediction results for all other datasets are given in Table~\ref{tab:link_pred2}. In the upper part of the table, we display the link prediction results for UMLS, KINSHIP, and NELL-995-h100. On UMLS, \approach{} + GS showed the second-best MRR, trailing behind \approach{} + LES. MuRP, MuRE and RotH achieved the highest performance for Hits at 1, 3 and 10 respectively. Notably, \approach{} + GSDC performed less effectively than \approach{}+LES only on UMLS data.

For the KINSHIP dataset, nearly all variations of our approach---except for \approach{} + VSP--- find the embedding space $Cl_{0,1,0}(\mathbb{R}^8)$ to be the most fitting. Due to the relationship between $Cl_{0,1,0}$, $Cl_{0,1}$, and $\mathbb{C}$ (refer to Table \ref{tab:clifford_subfields}), the performances of \approach{} + GSDC, \approach{} + GS, and \keci{} were similar. However, the scaling effect used by \keci{} \cite{demir2023clifford} resulted in slightly superior performance, making ComplEx the second-best. Similar observations can be made for the NELL-995-h50 dataset in the bottom part of the table, where \approach{} + GSDC performed embeddings into $Cl_{13,2,0}(\mathbb{R})$ which is theoretically isomorphic to $Cl_{13,2}(\mathbb{R})$, resulting in similar performance to \keci{}. Surprisingly, we remark that this is the only dataset where \approach{} + GS performed less than \approach{} + VSP, showing very close performance to Complex and  better results than DistMult, OMult, QMult and DualE.

Moving on to NELL-995-h100, \approach{} + GSDC achieved the highest performance, surpassing \keci{} by over $6\%$ in MRR. The second-best results were shared between \approach{} + LES and \approach{} + GS, which had similar performances.

On FB15k-237, \approach{} + GSDC demonstrated superior performance across all metrics. Remarkably, its Mean Reciprocal Rank (MRR) score is, on average, $40\%$ better than that of other approaches. While \approach{} + LES and \keci{} exhibited identical performance in Hits at 1 and MRR, \approach{} + LES took second place overall. This is attributed to its outperformance of \keci{} by more than $1\%$ in Hits at 3, despite \keci{} being only $0.2\%$ better in Hits at 10. Additionally, it's noteworthy that this is the only dataset in the study where DistMult outperformed one variant of \approach{} (\approach{} + VSP) across all metrics. This suggests that nilpotent vectors help capture the embeddings better on this dataset.


\section{Conclusion and Future Work}

In this paper, we introduce {\approach}, the first model specifically designed for embeddings in degenerate Clifford algebras. As a result, $\approach{}$ emerges as the most comprehensive divisional algebra in the literature for computing such embeddings.

We implemented four variants of \approach{} for optimal parameter search. The first two variants, {\approach}+LES and {\approach}+GSDC, considered as baselines use exhaustive searches and serve as upper bounds for our approach, yielding the best possible results. The third variant, \approach{}+GS, optimizes parameters within a subdomain of the parameter space, while the fourth variant, \approach{}+VSP, employs neural networks to predict parameters based on input data.

Evaluation in link prediction tasks across seven benchmark datasets consistently demonstrates the superiority of {\approach} over state-of-the-art models on all datasets. This underscores the potential of leveraging nilpotent vectors in Clifford algebras to enhance representation learning and inference capabilities in knowledge graphs.

Among the \approach{} variants, VSP generally yields slightly inferior results; however, its predicted results are consistently superior to most of the state-of-the-art models. We believe that this performance could be significantly improved with additional training data, thereby motivating future work.

Furthermore, in this study, we used only single base vectors (i.e. $1+p+q+r$ vectors) for entities and relations representation in $Cl_{p,q,r}(\mathbb{R})$ Clifford Algebras. Future work will explore the incorporation of multi-vectors, aiming to capture more intricate interactions in entities and relations by considering the full spectrum of the $2^{p+q+r}$ base vectors within a Clifford space $Cl_{p,q,r}(\mathbb{R})$.

\begin{ack}
    This project received funding from the European Union's Horizon Europe research and innovation programme through two grants (Marie Skłodowska-Curie grant No. 101073307 and grant No. 101070305). It was also supported by the "WHALE" project (LFN 1-04), funded by the Lamarr Fellow Network programme and the Ministry of Culture and Science of North Rhine-Westphalia (MKW NRW).
\end{ack}


\bibliography{mybibfile}

\begin{thebibliography}{28}
\providecommand{\natexlab}[1]{#1}
\providecommand{\url}[1]{\texttt{#1}}
\expandafter\ifx\csname urlstyle\endcsname\relax
  \providecommand{\doi}[1]{doi: #1}\else
  \providecommand{\doi}{doi: \begingroup \urlstyle{rm}\Url}\fi

\bibitem[Balazevic et~al.(2019)Balazevic, Allen, and
  Hospedales]{balazevic2019multi}
I.~Balazevic, C.~Allen, and T.~Hospedales.
\newblock Multi-relational poincar{\'e} graph embeddings.
\newblock \emph{Advances in Neural Information Processing Systems}, 32, 2019.

\bibitem[Bordes et~al.(2013)Bordes, Usunier, Garcia-Duran, Weston, and
  Yakhnenko]{bordes2013translating}
A.~Bordes, N.~Usunier, A.~Garcia-Duran, J.~Weston, and O.~Yakhnenko.
\newblock Translating embeddings for modeling multi-relational data.
\newblock \emph{Advances in neural information processing systems}, 26, 2013.

\bibitem[Cao et~al.(2021{\natexlab{a}})Cao, Xu, Yang, Cao, and
  Huang]{DBLP:conf/aaai/CaoX0CH21}
Z.~Cao, Q.~Xu, Z.~Yang, X.~Cao, and Q.~Huang.
\newblock Dual quaternion knowledge graph embeddings.
\newblock In \emph{Thirty-Fifth {AAAI} Conference on Artificial Intelligence,
  {AAAI} 2021, Thirty-Third Conference on Innovative Applications of Artificial
  Intelligence, {IAAI} 2021, The Eleventh Symposium on Educational Advances in
  Artificial Intelligence, {EAAI} 2021, Virtual Event, February 2-9, 2021},
  pages 6894--6902. {AAAI} Press, 2021{\natexlab{a}}.
\newblock \doi{10.1609/AAAI.V35I8.16850}.
\newblock URL \url{https://doi.org/10.1609/aaai.v35i8.16850}.

\bibitem[Cao et~al.(2021{\natexlab{b}})Cao, Xu, Yang, Cao, and
  Huang]{cao2021dual}
Z.~Cao, Q.~Xu, Z.~Yang, X.~Cao, and Q.~Huang.
\newblock Dual quaternion knowledge graph embeddings.
\newblock In \emph{Proceedings of the AAAI conference on artificial
  intelligence}, volume~35, pages 6894--6902, 2021{\natexlab{b}}.

\bibitem[Chami et~al.(2020)Chami, Wolf, Juan, Sala, Ravi, and
  R{\'e}]{chami2020low}
I.~Chami, A.~Wolf, D.-C. Juan, F.~Sala, S.~Ravi, and C.~R{\'e}.
\newblock Low-dimensional hyperbolic knowledge graph embeddings.
\newblock \emph{arXiv preprint arXiv:2005.00545}, 2020.

\bibitem[Chen et~al.(2020)Chen, Wang, Zhao, Cheng, Zhao, and
  Duan]{chen2020knowledge}
Z.~Chen, Y.~Wang, B.~Zhao, J.~Cheng, X.~Zhao, and Z.~Duan.
\newblock Knowledge graph completion: A review.
\newblock \emph{Ieee Access}, 8:\penalty0 192435--192456, 2020.

\bibitem[Clifford(2010)]{clifford2010clifford}
K.~Clifford.
\newblock Clifford algebra.
\newblock \emph{Quantum Algebra and Symmetry: Quantum Algebraic Topology,
  Quantum Field Theories and Higher Dimensional Algebra}, page 127, 2010.

\bibitem[Clifford(1882)]{clifford1882mathematical}
W.~K. Clifford.
\newblock \emph{Mathematical Papers by William Kingdon Clifford: Edited by
  Robert Tucker, with an introduction by HJ Stephen Smith}.
\newblock Macmillan and Company, 1882.

\bibitem[Demir and Ngonga~Ngomo(2023)]{demir2023clifford}
C.~Demir and A.-C. Ngonga~Ngomo.
\newblock Clifford embeddings--a generalized approach for embedding in normed
  algebras.
\newblock In \emph{Joint European Conference on Machine Learning and Knowledge
  Discovery in Databases}, pages 567--582. Springer, 2023.

\bibitem[Demir et~al.(2021)Demir, Moussallem, Heindorf, and
  Ngomo]{demir2021convolutional}
C.~Demir, D.~Moussallem, S.~Heindorf, and A.-C.~N. Ngomo.
\newblock Convolutional hypercomplex embeddings for link prediction.
\newblock In \emph{Asian Conference on Machine Learning}, pages 656--671. PMLR,
  2021.

\bibitem[Fu et~al.(2016)Fu, Zhang, and Li]{fu2016using}
R.~Fu, Z.~Zhang, and L.~Li.
\newblock Using lstm and gru neural network methods for traffic flow
  prediction.
\newblock In \emph{2016 31st Youth academic annual conference of Chinese
  association of automation (YAC)}, pages 324--328. IEEE, 2016.

\bibitem[Hogan et~al.(2022)Hogan, Blomqvist, Cochez, d'Amato, de~Melo,
  Gutierrez, Kirrane, Gayo, Navigli, Neumaier, Ngomo, Polleres, Rashid, Rula,
  Schmelzeisen, Sequeda, Staab, and
  Zimmermann]{DBLP:journals/csur/HoganBCdMGKGNNN21}
A.~Hogan, E.~Blomqvist, M.~Cochez, C.~d'Amato, G.~de~Melo, C.~Gutierrez,
  S.~Kirrane, J.~E.~L. Gayo, R.~Navigli, S.~Neumaier, A.~N. Ngomo, A.~Polleres,
  S.~M. Rashid, A.~Rula, L.~Schmelzeisen, J.~F. Sequeda, S.~Staab, and
  A.~Zimmermann.
\newblock Knowledge graphs.
\newblock \emph{{ACM} Comput. Surv.}, 54\penalty0 (4):\penalty0 71:1--71:37,
  2022.
\newblock \doi{10.1145/3447772}.
\newblock URL \url{https://doi.org/10.1145/3447772}.

\bibitem[Ji et~al.(2015)Ji, He, Xu, Liu, and Zhao]{ji2015knowledge}
G.~Ji, S.~He, L.~Xu, K.~Liu, and J.~Zhao.
\newblock Knowledge graph embedding via dynamic mapping matrix.
\newblock In \emph{Proceedings of the 53rd annual meeting of the association
  for computational linguistics and the 7th international joint conference on
  natural language processing (volume 1: Long papers)}, pages 687--696, 2015.

\bibitem[Ji et~al.(2022)Ji, Pan, Cambria, Marttinen, and
  Yu]{DBLP:journals/tnn/JiPCMY22}
S.~Ji, S.~Pan, E.~Cambria, P.~Marttinen, and P.~S. Yu.
\newblock A survey on knowledge graphs: Representation, acquisition, and
  applications.
\newblock \emph{{IEEE} Trans. Neural Networks Learn. Syst.}, 33\penalty0
  (2):\penalty0 494--514, 2022.
\newblock \doi{10.1109/TNNLS.2021.3070843}.
\newblock URL \url{https://doi.org/10.1109/TNNLS.2021.3070843}.

\bibitem[Jia(2013)]{jia2013dual}
Y.-B. Jia.
\newblock Dual quaternions.
\newblock \emph{Iowa State University: Ames, IA, USA}, 2013.

\bibitem[KAMDEM~TEYOU et~al.(2024)KAMDEM~TEYOU, Demir, and
  Ngonga~Ngomo]{kamdem_teyou_2024_13341499}
L.~M. KAMDEM~TEYOU, C.~Demir, and A.-C. Ngonga~Ngomo.
\newblock {Appendix: Embedding Knowledge Graphs in Degenerate Clifford
  Algebras}, Aug. 2024.
\newblock URL \url{https://doi.org/10.5281/zenodo.13341499}.

\bibitem[Keller and Ochsenius(2012)]{keller2012structure}
H.~A. Keller and H.~Ochsenius.
\newblock The structure of norm clifford algebras.
\newblock \emph{Mathematica Slovaca}, 62\penalty0 (6):\penalty0 1105--1120,
  2012.

\bibitem[Lin et~al.(2015)Lin, Liu, Sun, Liu, and Zhu]{lin2015learning}
Y.~Lin, Z.~Liu, M.~Sun, Y.~Liu, and X.~Zhu.
\newblock Learning entity and relation embeddings for knowledge graph
  completion.
\newblock In \emph{Proceedings of the AAAI conference on artificial
  intelligence}, volume~29, 2015.

\bibitem[Nickel et~al.(2011)Nickel, Tresp, Kriegel, et~al.]{nickel2011three}
M.~Nickel, V.~Tresp, H.-P. Kriegel, et~al.
\newblock A three-way model for collective learning on multi-relational data.
\newblock In \emph{Icml}, volume~11, pages 3104482--3104584, 2011.

\bibitem[Singhal(2012)]{Singhal2012}
A.~Singhal.
\newblock Introducing the knowledge graph: Things, not strings.
\newblock Google Blog, 2012.
\newblock Retrieved from
  \url{https://www.blog.google/products/search/introducing-knowledge-graph-things-not/}.

\bibitem[Sun et~al.(2019)Sun, Deng, Nie, and Tang]{sun2019rotate}
Z.~Sun, Z.-H. Deng, J.-Y. Nie, and J.~Tang.
\newblock Rotate: Knowledge graph embedding by relational rotation in complex
  space.
\newblock \emph{arXiv preprint arXiv:1902.10197}, 2019.

\bibitem[Trouillon et~al.(2016{\natexlab{a}})Trouillon, Welbl, Riedel,
  Gaussier, and Bouchard]{DBLP:conf/icml/TrouillonWRGB16}
T.~Trouillon, J.~Welbl, S.~Riedel, {\'{E}}.~Gaussier, and G.~Bouchard.
\newblock Complex embeddings for simple link prediction.
\newblock In M.~Balcan and K.~Q. Weinberger, editors, \emph{Proceedings of the
  33nd International Conference on Machine Learning, {ICML} 2016, New York
  City, NY, USA, June 19-24, 2016}, volume~48 of \emph{{JMLR} Workshop and
  Conference Proceedings}, pages 2071--2080. JMLR.org, 2016{\natexlab{a}}.
\newblock URL \url{http://proceedings.mlr.press/v48/trouillon16.html}.

\bibitem[Trouillon et~al.(2016{\natexlab{b}})Trouillon, Welbl, Riedel,
  Gaussier, and Bouchard]{trouillon2016complex}
T.~Trouillon, J.~Welbl, S.~Riedel, {\'E}.~Gaussier, and G.~Bouchard.
\newblock Complex embeddings for simple link prediction.
\newblock In \emph{International conference on machine learning}, pages
  2071--2080. PMLR, 2016{\natexlab{b}}.

\bibitem[Wang et~al.(2017)Wang, Mao, Wang, and
  Guo]{DBLP:journals/tkde/WangMWG17}
Q.~Wang, Z.~Mao, B.~Wang, and L.~Guo.
\newblock Knowledge graph embedding: {A} survey of approaches and applications.
\newblock \emph{{IEEE} Trans. Knowl. Data Eng.}, 29\penalty0 (12):\penalty0
  2724--2743, 2017.
\newblock \doi{10.1109/TKDE.2017.2754499}.
\newblock URL \url{https://doi.org/10.1109/TKDE.2017.2754499}.

\bibitem[Wang et~al.(2014)Wang, Zhang, Feng, and Chen]{wang2014knowledge}
Z.~Wang, J.~Zhang, J.~Feng, and Z.~Chen.
\newblock Knowledge graph embedding by translating on hyperplanes.
\newblock In \emph{Proceedings of the AAAI conference on artificial
  intelligence}, volume~28, 2014.

\bibitem[Xu et~al.(2020)Xu, Nayyeri, Chen, and Lehmann]{xu2020knowledge}
C.~Xu, M.~Nayyeri, Y.-Y. Chen, and J.~Lehmann.
\newblock Knowledge graph embeddings in geometric algebras.
\newblock \emph{arXiv preprint arXiv:2010.00989}, 2020.

\bibitem[Yang et~al.(2014)Yang, Yih, He, Gao, and Deng]{yang2014embedding}
B.~Yang, W.-t. Yih, X.~He, J.~Gao, and L.~Deng.
\newblock Embedding entities and relations for learning and inference in
  knowledge bases.
\newblock \emph{arXiv preprint arXiv:1412.6575}, 2014.

\bibitem[Zhang et~al.(2019)Zhang, Tay, Yao, and Liu]{zhang2019quaternion}
S.~Zhang, Y.~Tay, L.~Yao, and Q.~Liu.
\newblock Quaternion knowledge graph embeddings.
\newblock \emph{Advances in neural information processing systems}, 32, 2019.

\end{thebibliography}

\end{document}